\documentclass[acmtog,authorversion,nonacm]{acmart}
\usepackage{multirow}

\usepackage{booktabs} 

\usepackage[ruled]{algorithm2e} 

\SetAlFnt{\small}
\SetAlCapFnt{\small}
\SetAlCapNameFnt{\small}
\SetAlCapHSkip{0pt}

\begin{document}
\title{ClothCombo: Modeling Inter-Cloth Interaction for Draping Multi-Layered Clothes}

\author{Dohae Lee}
\orcid{0000-0003-4593-6614}
\affiliation{%
 \institution{Yonsei University}
 \department{Dept. of Computer Science}
 \streetaddress{Yonsei-Ro 50}
 \state{Seodaemungu}
 \city{Seoul}
 \country{Rep. of Korea}}
\email{dlehgo1414@yonsei.ac.kr}

\author{Hyun Kang}
\orcid{0009-0002-7571-9258}
\affiliation{%
 \institution{Yonsei University}
 \department{Dept. of Computer Science}
 \streetaddress{Yonsei-Ro 50}
 \state{Seodaemungu}
 \city{Seoul}
 \country{Rep. of Korea}}
\email{kdkh1125@yonsei.ac.kr}

\author{In-Kwon Lee}
\orcid{0000-0002-1534-1882}
\affiliation{%
 \institution{Yonsei University}
 \department{Dept. of Computer Science}
 \streetaddress{Yonsei-Ro 50}
 \state{Seodaemungu}
 \city{Seoul}
 \country{Rep. of Korea}}
\email{iklee@yonsei.ac.kr}

\renewcommand\shortauthors{Lee et al}

\begin{abstract}
We present ClothCombo, a pipeline to drape arbitrary combinations of clothes on 3D human models with varying body shapes and poses. 
While existing learning-based approaches for draping clothes have shown promising results, multi-layered clothing remains challenging as it is non-trivial to model inter-cloth interaction. 
To this end, our method utilizes a GNN-based network to efficiently model the interaction between clothes in different layers, thus enabling multi-layered clothing. 
Specifically, we first create feature embedding for each cloth using a topology-agnostic network. 
Then, the draping network deforms all clothes to fit the target body shape and pose without considering inter-cloth interaction. 
Lastly, the untangling network predicts the per-vertex displacements in a way that resolves interpenetration between clothes. 
In experiments, the proposed model demonstrates strong performance in complex multi-layered scenarios. Being agnostic to cloth topology, our method can be readily used for layered virtual try-on of real clothes in diverse poses and combinations of clothes. 

\end{abstract}

%
%

%
%

\begin{teaserfigure}
  \includegraphics[width=1.0\linewidth]{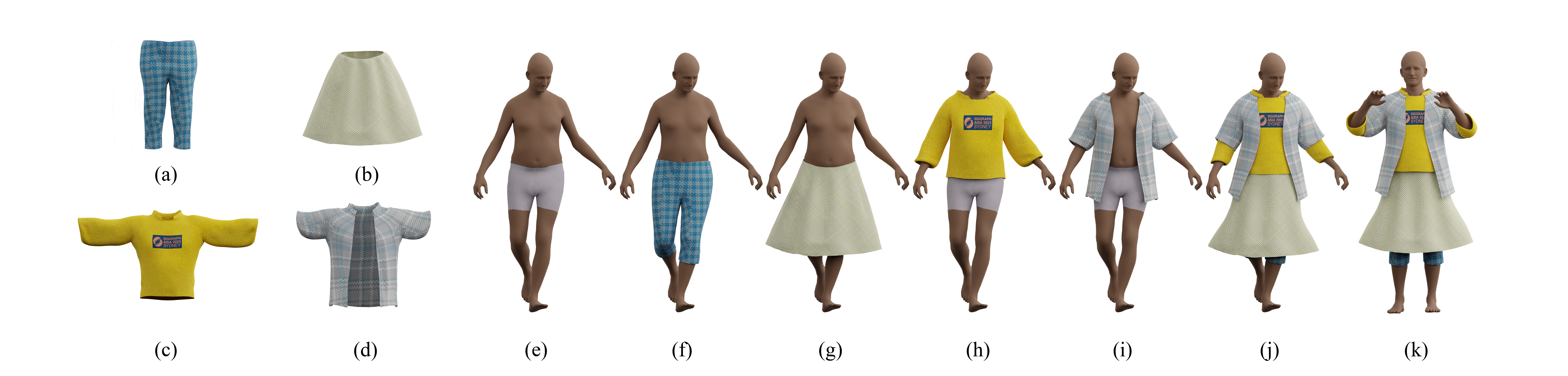}
  \caption{(a-d) The target clothes, (e) the target body, (f-i) the results of draping each individual clothes, (j) the result of draping multi-layered garments in order of (f) to (i), and (k) the result of draping the clothes in a different pose.}
  \label{fig:teaser}
\end{teaserfigure}

\maketitle

\section{Introduction}

Cloth simulation is a technique used to animate the movement of cloth in a computer program, mimicking the behavior of real cloth in the virtual world. 
It is a long-standing area of interest in the computer graphics community with its diverse applications in animation, gaming, virtual reality, etc \cite{Nealen, baraff_large_step, ko_research}.
Traditional methods are based on physically-based simulation, which applies physical laws and mathematical algorithms to a virtual model of the cloth.
While realistic results can be achieved, a large amount of memory and computational resources are required, limiting its deployment to real-time applications.
Learning-based methods have been proposed as alternatives to traditional cloth simulation, serving the purpose of draping clothes on a wide range of body shapes and poses in real-time \cite{snug, TailorNet, PBNS, AgentDress, sig2022}.
However, these methods typically require training for each clothes, making it difficult to scale with countless number of clothes in the world. 
Several studies have introduced methods to handle clothes with arbitrary topology, yet they are not generalizable to address complex multi-layered scenarios in which several clothes are intricately tangled \cite{DeePSD, DrapeNet, GarNet++, hood, deep_detail}.
Recently, research on untangling clothes of multiple layers has been proposed, but it was only possible in the T-pose, required learning in the preprocess stage for each garment, and was difficult to apply to complex scenarios \cite{ULNeF}.

In this paper, we propose a novel approach \emph{ClothCombo} to drape arbitrary combinations of clothes on a wide range of body shapes and poses. 
Our proposed framework consists of three sub-networks: the clothes embedding network, the single clothing draping network, and the untangling network. 
The clothes embedding network represents the geometries of clothes into a unified representation using a topology-agnostic network. 
The single clothing draping network deforms the clothes to fit the target body shape and pose. It does not yet consider the inter-clothes interaction between different clothes layers. 
The core of our framework is the untangling network, which deforms the clothes in a way that prevents penetrations between clothes in each layer.
It uses a graph-based modeling for inter-clothes interaction, in which each garment is represented as a node and interaction between the garments as a message passing through the edge.
The use of compact representation for each garment and a simple yet effective GNN-based network for inter-clothes interaction is generalizable to complex multi-layered scenarios as shown in Figure \ref{fig:teaser}.
The entire training process is unsupervised, optimized solely by physics constraint loss terms. This eliminates the need to generate extensive synthetic data with varied clothing layers, orders, body shapes, and poses, which is notably challenging.

In our experiments, we evaluate our framework's performance on synthetic and real-world 3D clothes. We demonstrate that our proposed method can simulate the draping of complex multi-layered clothing in a more physically plausible way compared to previous methods. 
Additionally, we conduct an ablation study to validate the importance of each component of our proposed framework in achieving these results.

Our main contributions are summarized as:
\begin{itemize}
    \item The capability to drape clothes with arbitrary topology using a topology-agnostic clothes embedding network.
    \item The simulation of complex multi-layered clothing on diverse body shapes and poses through the use of a GNN-based inter-clothes interaction modeling method.
    \item The applicability in a downstream task such as layered 3D virtual try-on applications.
\end{itemize}

\section{Related Work}

\subsection{Physics-based Cloth Simulation}
The field of cloth simulation has been of significant interest within the computer graphics community for many years  \cite{Nealen, baraff_large_step, ko_research}. 
The use of advanced techniques such as integration methods \cite{integration1, integration2, integration3, elastically, integration4, integration5}, strain limiting \cite{strain1, strain2, strain3, strain4}, and yarn-level simulation \cite{yarn1, yarn2, yarn3} have resulted in highly realistic simulations.
However, these methods require high computational time, which can restrict their use in real-time applications, particularly for high-resolution cloth geometry.
To address this issue, many approaches have been proposed such as parallel computation \cite{parallel1, parallel2}, coarse-to-fine approaches \cite{coarse2fine1, coarse2fine2,coarse2fine3,coarse2fine4}, position-based dynamics \cite{position1, position2}, projective dynamics \cite{projective1, projective2}, and example-based methods \cite{example1, example2} among others \cite{other1, other2}.
However, these methods still may not provide real-time performance or result in loss of fine details.

\begin{figure*}[t]
  \includegraphics[width=1.0\linewidth]{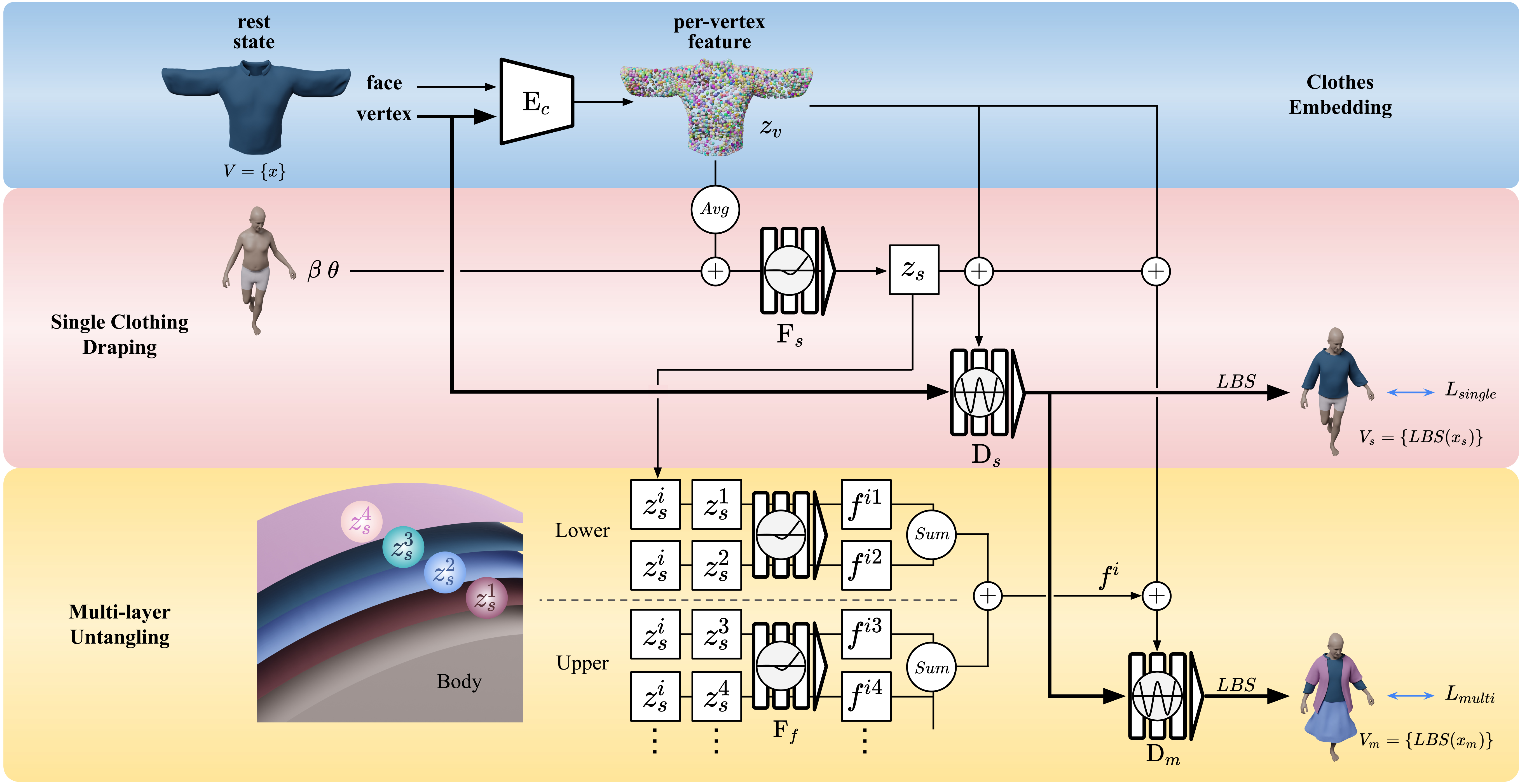}
  \caption{Model overview.}
  \label{fig:model_overview}
\end{figure*}

\subsection{Learning-based Cloth Simulation}

Recently, the advancements in neural networks have greatly improved physics-based simulations in various fields, such as smoke simulation \cite{smoke}, fluid dynamics \cite{fluid}, and rigid body dynamics \cite{rigid}. In the field of cloth simulation, ongoing research is focusing on various topics, including simulation \cite{clothsim1}, reconstruction \cite{clothrecon1, clothrecon2, clothrecon3, clothrecon4, clothrecon5}, generation \cite{clothgen1, clothgen2, clothgen3, clothgen4, clothgen5, clothgen6}, animation of real-world cloth \cite{real1, real2}, and the generation of sewing patterns \cite{pattern1, pattern2, pattern3}.

Recently, there has been active research on methods for dressing avatars represented by parametric human models such as SCAPE \cite{SCAPE} and SMPL \cite{SMPL}. These methods are computationally more efficient than physics-based simulations and make it more convenient for animators to dress avatars by eliminating the need for manual alignment of the body and clothes. DRAPE \cite{drape} was the first method proposed for modeling clothes deformation based on pose and shape. Subsequently, other methods such as DeepWrinkles \cite{DeepWrinkles}, which express global shape deformation as geometry and high-frequency details as a normal map, and the method proposed in \cite{santesteban2019}, which separately models global deformations caused by the shape and fine wrinkles caused by the shape and pose, were introduced. Additionally, GarNet \cite{GarNet} is a method proposed for quickly draping roughly aligned clothes on the body, while TailorNet \cite{TailorNet} enables draping based on pose, shape, and clothes style in canonical shape and pose space.
AgentDress \cite{AgentDress} enables virtual try-on in real-time on the CPU using an example database, and DeePSD \cite{DeePSD} predicts the shape key and blend weights from a given clothes template to enable efficient simulation and improves generalization performance by using physics-based loss. In addition, fully self-supervised training methods, such as \cite{snug, PBNS}, have been proposed, which do not require a large amount of data generated using physics-based simulation. However, most studies focus on template-based clothes rather than entirely different topologies. 
To address this limitation, several methods have demonstrated its applicability to clothes in arbitrary topology by encoding it as a point-cloud \cite{GarNet++, DeePSD, DrapeNet}, a graph \cite{hood, physgraph}, a mesh \cite{wang2019learning}, or an UV-map \cite{deep_detail}.

However, existing methods can only handle single clothes, or those with limited areas of overlap \cite{DrapeNet} or require an initially untangled template \cite{DeePSD}. A research has shown its applicability in multi-layered clothing, but it requires training multiple networks sequentially which limits its scalability with the combinatorial nature of layered clothing \cite{motion_guided}. Recently, a method called ULNeF \cite{ULNeF} was proposed for quickly untangling layered clothes. However, ULNeF requires learning to obtain the implicit field for each garment with diverse shapes as a preprocess and is challenging to apply to complex multi-layered scenarios. Additionally, ULNeF can only be applied in the T-pose. Our proposed method \emph{ClothCombo} addresses these limitations by allowing for realistic draping of arbitrary combination of clothes on avatars with diverse poses and shapes that is generalizable to complex multi-layer scenarios.

\section{Method}
\label{sec:method}

The proposed method, referred to as ClothCombo, aims to drape multi-layer garments with arbitrary topologies on a wide range of human body shapes and poses.
Figure \ref{fig:model_overview} provides an overview of our model, wherein the top blue section, the middle red section, and the last yellow section represent the clothes embedding phase (detailed in Section \ref{sec:clothes_embedding}), the single garment draping phase (detailed in Section \ref{sec:SGND}), and the untangling phase (detailed in Section \ref{sec:untangling}), respectively.
We will begin by providing an explanation of our human body representation in the following section (Section \ref{sec:body_representation}), and subsequently describe each component of our framework in the order they are presented.
Finally, in Section \ref{sec:training}, we will explain the training scheme and loss functions utilized in our method.

\subsection{Human Body Representation}
\label{sec:body_representation}

The Skinned Multi-Person Linear model (SMPL) \cite{SMPL} is a parametric body model that has been trained on thousands of 3D human scan data, allowing it to represent various 3D human bodies through shape ($\beta$) and pose ($\theta$) parameters:
\begin{equation}
\label{eq:SMPL1}
    H(\beta, \theta) = W(T_P(\beta, \theta), J(\beta), \theta, \mathcal{W}),
\end{equation}
\begin{equation}
    T_P(\beta, \theta) = \bar{T}+B_S(\beta)+B_P(\theta),
\end{equation}
where the final human body mesh $H(\beta, \theta)$ is determined by the Linear Blend Skinning (LBS) function $W$, using the positions of joints $J(\beta)$, and the skinning weights $\mathcal{W}$. The deformation of the template body mesh $\bar{T}$, represented as $T_P(\beta, \theta)$, is a result of both shape-dependent blend-shapes $B_S(\beta)$, and pose-dependent deformation $B_P(\theta)$. This work utilizes SMPL-X \cite{SMPL-X}, an extended version of SMPL \cite{SMPL}, which incorporates the hand representation model MANO \cite{MANO} and the facial shape and expression representation model FLAME \cite{FLAME}.

\subsection{Clothes Embedding Phase}
\label{sec:clothes_embedding}

The initial phase of the process is the clothes embedding, in which we aim to represent each garment in a topology-agnostic manner, as illustrated in the upper blue section of Figure \ref{fig:model_overview}. 
Among the candidates that satisfy this condition, including the point-based methods \cite{pointcnn, pointnet, pointnet++, kpconv, dgcnn} and the mesh-based methods \cite{meshcnn, hodgenet, diffusionnet, fieldconv, meshwalker}, DiffusionNet \cite{diffusionnet} is used as the backbone of our embedding network. 
As the input to our network, each garment is roughly aligned with the mean shape, T-posed body (zero-body). We say this garment is in the ``rest state''. 
Capturing geometric attribute of the garment in the rest state, the clothes embedding network generates the per-vertex clothes embedding, $z_v$, along with the global clothes embedding, $\bar{z}_v$, as the mass ($m_v$) weighted average of all vertex features \cite{diffusionnet}:
\begin{equation}
    {z_v}=\mathrm{E_c}(V, F),
\end{equation}
\begin{equation}
    {\bar{z}_v}=Avg(z_v m_v),
\end{equation}
where $\mathrm{E_c}$ is the embedding network, and $V=\{x_i\}_{i=1}^{N}$ and $F$ are the set of vertex coordinates and the set of faces of clothes in the rest state, respectively.

\subsection{Single Clothing Draping Phase}
\label{sec:SGND}

The single clothing draping network is designed to drape a single piece of clothing on a human body model.
Specifically, the network predicts the displacement of each vertex in the canonical space, which represents the deformation caused by the underlying body shape and pose.
The displacement is added to the clothes in the rest state, which then is transformed into the posed space via linear blend skinning (LBS) with the learned skinning weights from the skinning network (Section \ref{sec:skinning_network}).

We first derive the single draping embedding $z_s$ from the global clothes embedding $\bar{z}_v$, the body shape $\beta$ and the pose $\theta$:
\begin{equation}
z_s = \mathrm{F}_s(\bar{z}_v, \beta, \theta),
\end{equation}
where $\mathrm{F}_s$ is a learnable function that encodes the latent vector of clothes in the target body shape and pose.
Subsequently, we predict the per-vertex displacement $\Delta x_s$, which models the deformation due to the shape and pose of the body:
\begin{equation}
\Delta x_s = \mathrm{D_s}(x, z_s, {z}_v),
\end{equation}
where $\mathrm{D_s}$ is a function that predicts the displacement for each vertex in the rest state $x$. It is implemented with a modulated SIREN \cite{modulatedSIREN}.
Finally, the set of vertices $V_s$ draped on the target body is given by:
\begin{equation}
V_s = \left \{W \left (x_s, J(\beta), \theta, \mathcal{W}_x + \Delta\mathcal{W}_x \right )\right \},
\end{equation}
\begin{equation}
x_s = x + \Delta x_s,
\end{equation}
where the deformed vertices, $x_s$, are mapped to the posed-space by the optimized skinning weights from the skinning network, $\mathcal{W}_x + \Delta\mathcal{W}_x$.

\subsubsection{Skinning Network}
\label{sec:skinning_network}

The skinning weight of each vertex in the clothes, represented by $\mathcal{W}_x$, is initially determined by the skinning weight of the nearest vertex on the zero-body mesh. However, to enhance the precision of the skinning weights, we employ a skinning network, denoted as $\mathrm{S}$, to calculate the necessary adjustments to the skinning weights with respect to the target body shape and pose. Formally, we take the coordinate of each vertex $x$, the single draping embedding $z_s$, and the per-vertex clothes embedding $z_v$:
\begin{equation}
    \Delta \mathcal{W}_x = \mathrm{S}(x, z_s, z_v).
\end{equation}
The calculated adjustments $\Delta \mathcal{W}_x$ are then added to the initial skinning weight of the vertex, resulting in an optimized skinning weight, $\mathcal{W}_x + \Delta \mathcal{W}_x$.

\subsection{Untangling Phase}
\label{sec:untangling}

The previous phase involves draping a single garment individually on the target body, generating the single draping embedding $z_s$ for each garment. This embedding contains information about how the garment deforms in accordance with the underlying body.
To extend this concept to multi-layered draping, we can conceptualize the garments across all layers as a single, fully-connected graph.
Each node in the graph represents a garment, and each edge represents an interaction between the garments. 
To model the forces exchanged between garments and account for their interactions, we design a network based on Graph Neural Network (GNN) \cite{discovering}, utilizing this graph structure.

Let $M$ denote the number of layers, or the number of garments that are layered. We denote the set of vertex coordinates of the $i$-th garment $(i = 1, 2, ..., M)$ in layer $i$ of the rest state by $V^i=\{x_1^i, ..., x_N^i \}$, where $x^i_j$ represents the $j$-th vertex coordinates in total $N$ vertices of $i$-th garment. For the brevity of notation, we will omit the indices $i$ and $j$ and instead denote the set of vertex coordinates as $V=\{x\}$. We assume that the layer number $i$ corresponds to the sequence in which the garments are draped over the human body. For instance, the garment on layer 1 is draped onto the body first, followed by the garment on layer 2, and so forth.
To model the interactions between clothes in different layers and prevent intersections between them, we utilize the function $\mathrm{F_f}$, which takes the single draping embedding, $z_s^i$ and $z_s^j$ as input, of the $i$-th and $j$-th clothes, respectively:
\begin{equation}
	f^{ij} = \mathrm{F_f} \left ( z_s^i, z_s^j \right ),
\end{equation}
where $f^{ij}$ represents the force between these two layers. 
The net force $f^i$ acting on the $i$-th clothes is then calculated as the concatenation of the sum of forces from the lower and higher layers:
\begin{equation}
	f^i = \sum_{i>j}f^{ij} \oplus \sum_{i<j}f^{ij},
\end{equation}
where $\oplus$ denotes the concatenation.
The vertex displacements of the $i$-th garment, $\Delta x_m^i$, represent the deformation caused by the interaction with other garment layers:
\begin{equation}
	\Delta x_m^i = \mathrm{D_m}
	\left (x^i_s, f^i, z^i_s, z^i_v \right ),
\end{equation}
where $x^i_s$ is $x_s$ of the $i$-th garment.
Since the untangling process so far has been done in the T-pose space, it is necessary to transform the vertices of the clothes to the posed space using Linear Blend Skinning (LBS) technique to obtain the final multi-layered clothing vertex positions $V_m$:
\begin{equation}
	V_m = \{LBS(x_m)\} = \left \{ W
	\left ( x_m, J(\beta), \theta, \mathcal{W}_x + 		\Delta\mathcal{W}_x \right )\right \}.
\end{equation}

\subsection{Training}
\label{sec:training}

We first pre-train the clothes embedding network, the single clothing draping network, and the skinning network jointly. Then, we proceed to train all networks in an integrated, end-to-end manner. Our method employs a progressive learning strategy, in which the transformation from T-pose space to posed-space is incrementally implemented.

\subsubsection{Training Single Clothing Draping Network}
\label{sec:training_sgdn}

In order to ensure that the single garment conforms to the body and deform realistically, we employ a set of loss terms. Some of these loss terms, specifically the strain loss $L_{s}$, bending loss $L_{b}$, gravity loss $L_{g}$, and collision loss $L_c$, have been previously proposed in the literature, specifically in SNUG \cite{snug}. Additionally, we introduce several novel loss terms in this work, such as the repulsive loss $L_{r}$ to prevent self-intersection, and the holding loss $L_{h}$ to prevent the garment from slipping away from the body, particularly for bottom garments. For clarity, the set of $N$ vertex coordinates of the garment in its rest state is denoted as 
$V=\{x_j\}_{j=1}^{N}$, and the set of coordinates of deformed vertices by the single garment draping network is denoted as 
$V_s=\{\hat{x}_j\}_{j=1}^{N}$.

\textbf{Strain loss} is employed to ensure that the shape of the polygons in the deformed garment remains as close as possible to their rest state. This loss term is based on the Saint Venant Kirchhoff (StVK) elastic material model, which is formulated as follows:
\begin{equation}
\label{eq:strain}
    L_s = \sum_{triangles}\Big(\frac{\lambda_{L}}{2} tr(T_G)^2+\mu_{L}tr(T_G^2)\Big)\mathcal{V},
\end{equation}
where $\lambda_{L}$ and $\mu_L$ represent the Lamé constants, and $\mathcal{V}$ is the volume of each triangle, $T_G=\frac{1}{2}(F^\top F-\mathbf{I})$, and $F=\frac{\partial \hat{x}}{\partial \tilde{x}}$ denotes the deformation gradient, with the deformed position denoted by $\hat{x}$, and the rest position represented by ${x}$. Additionally, $\mathbf{I}$ represents the identity matrix. In our experiments, we set $\lambda_{L}$ to $4.44e$+$4$ and $\mu_L$ to $2.36e$+$4$.

\textbf{Bending loss} is employed to minimize the bending of adjacent polygons, which is formulated as follows:
\begin{equation}
L_b = \sum_{edges}\frac{k_{b}}{2}\alpha^2,
\end{equation}
where $k_b$ represents the bending stiffness, and $\alpha$ is the radian between two adjacent polygons. 
This loss term effectively constrains the deformation of the garment to prevent excessive bending of the polygons, ensuring a more realistic appearance.

\textbf{Gravity loss} is utilized to ensure that the vertices of the garment fall naturally under the influence of gravity. For all deformed vertices, denoted by $\hat{x}$, it is defined as follows:
\begin{equation}
L_g = \sum_{\hat{x}\in V_s}-m\mathbf{g}^\top \hat{x},
\end{equation}
where $m$ represents the mass of the vertex, and $\mathbf{g}$ denotes the acceleration due to gravity. 
This loss term effectively simulates the effects of gravity on the garment, ensuring that it behaves in a manner consistent with physical reality.

\textbf{Collision loss} is employed to ensure that the vertices of the garment do not intersect with the body mesh. This is achieved through the following formulation:
\begin{equation}
\label{eq:col}
    L_c = max(\epsilon - d(\hat{x}), 0)^3,
\end{equation}
where $d(\hat{x})=(\hat{x} - x_b)^\top n_b$ represents the distance between the closest body vertex, denoted by $x_b$, and the deformed vertex, $\hat{x}$, with $n_b$ being the normal vector of the closest body vertex. Additionally, $\epsilon$ represents the given minimal distance between the body and the garment. 
This loss term effectively constrains the deformation of the garment to prevent intersection with the body mesh, resulting in a more realistic and physically plausible outcome.

\textbf{Repulsive loss} is utilized to prevent self-intersection within the garment, which is formulated as follows:
\begin{equation}
    L_r = \sum_i^N{\sum_{j=i+1}^N{-log 
    \left ( (\hat{x_i} - \hat{x_j})^2 \right )\mathcal{A}_{ij}}},
\end{equation}
where $\hat{x_i}, \hat{x_j} \in {V_s}$, and $\mathcal{A}_{ij}=1$  if both conditions are met: (1) $\hat{x_i}$ and $\hat{x_j}$ are not connected by an edge and (2) the distance between $\hat{x_i}$ and $\hat{x_j}$ is less than a pre-defined threshold (which we set to 0.1m). If either condition is not met, $\mathcal{A}_{ij}=0$. 
This loss term effectively prevents self-intersection in the garment by constraining the relative distances between non-adjacent vertices not to be too close, resulting in a more physically plausible outcome.

\textbf{Holding loss} is exclusively applied to trousers, skirts or strapless tops to prevent them from slipping away from the body. This is achieved through the following formulation:
\begin{equation}
    L_h = \sum_i^N y(\Delta \hat{x_i})^2H_i,
\end{equation}
where $y(.)$ denotes the $y$-coordinates, and $H_i = 1$ for the top 10\% vertices in the rest state, and $H_i=0$ for all other vertices. This loss term effectively constrains the deformation of the bottom garment, specifically targeting the vertices in the lowermost region, to prevent them from slipping away from the body, allowing the bottom garments to be trained effectively.
All of these loss terms are combined to form the final loss function $L_{single}$ for training the single garment draping network, which is formulated as follows:
\begin{equation}
    L_{single} = \lambda_s L_s + \lambda_b L_b + \lambda_g L_g + \lambda_c L_c + \lambda_r L_r + \lambda_h L_h,
\end{equation}
where the loss weights, $\lambda_s$ and $\lambda_g$ are set to 1, $\lambda_b$ is set to 5, $\lambda_c$ is set to 250, $\lambda_r$ is set to 0.001, and $\lambda_h$ is set to 100 in our experiments. This loss term effectively combines all of the aforementioned loss terms to optimize the performance of the single garment draping network, enabling it to drape the various garments on diverse body shapes and poses.

\subsubsection{Training Untangling Network}
\label{sec:training_untangling}

To train our untangling network, we incorporate additional loss terms to prevent intersections between garments, specifically the multi-collision loss ($L_{mc}$), and the distance loss ($L_{d}$).
These loss terms effectively constrain the relative positions and distances between multiple garments, preventing intersections and ensuring a physically plausible outcome.
In our notation, the vertices in the $i$-th final deformed garment are denoted as $V^i_m=\big\{x^i_m\big\}$.

\textbf{Multi-collision loss}, in the scenario where there are a total of $M$ layers of garments, denoted as $L_{mc}^{ij}$, between the $i$-th and $j$-th garments, where the $j$-th garment is situated above the $i$-th garment ($i<j$), can be defined as follows:
\begin{equation} 
    L_{mc}^{ij} = \sum_{x^j \in V^j_m} max 
    \left ( \epsilon - d_g^i({x}^j), 0 \right )^3,
\end{equation}
where $d_g^i({x}^j)=({x}^j - x_g^i)^\top n_g^i$, with $x_g^i$ representing the coordinates of the closest vertex in the $i$-th final deformed garment $V_m^i$ from ${x}^j$, and $n_g^i$ representing the normal of that vertex.
It is designed to prevent the interpenetration between the garments in the different layers.

\textbf{Distance loss} is employed to ensure that the lower-layer garment is always positioned closer to the underlying body than the upper garment. The Distance loss between the $i$-th and $j$-th garments ($i<j$) is defined as:
\begin{equation}
    L^{ij}_d = \sum_{x^i \in V^i_m}\sum_{x^j \in V^j_m}{\Big(max\big(d(x^i)-d(x^j), 0\big)^3\mathfrak{B}_{{x^i}{x^j}} \Big) },
\end{equation}
where $d(x)$ is defined in Equation (\ref{eq:col}), and $\mathfrak{B}_{{x^i}{x^j}}$ is set to $1$ if both conditions are met: (1) the closest body vertex is the same in $x^i$ and $x^j$, and (2) both $d(x^i)$ and $d(x^j)$ are less than a threshold of $0.1$ m; otherwise, it is set to $0$.

Finally, the comprehensive loss term for training the untangling network, denoted as $L_{multi}$, is defined as follows:
\begin{equation}
    L_{multi} = L_{single} + \sum_{i=1}^M\sum_{j=i+1}^M{\left( \lambda_{mc} L_{mc}^{ij} + \lambda_{d} L_d^{ij} \right)},
\end{equation}
where $L_{single}$ is the loss term as described in Section \ref{sec:training_sgdn}, and the loss weights $\lambda_{mc}$ is set to 250, and $\lambda_{d}$ is set to 25000 in our experiments.

\section{Experiments}
\label{sec:experiments}

\subsection{Dataset}

\begin{figure}[t]
  \includegraphics[width=1.0\linewidth]{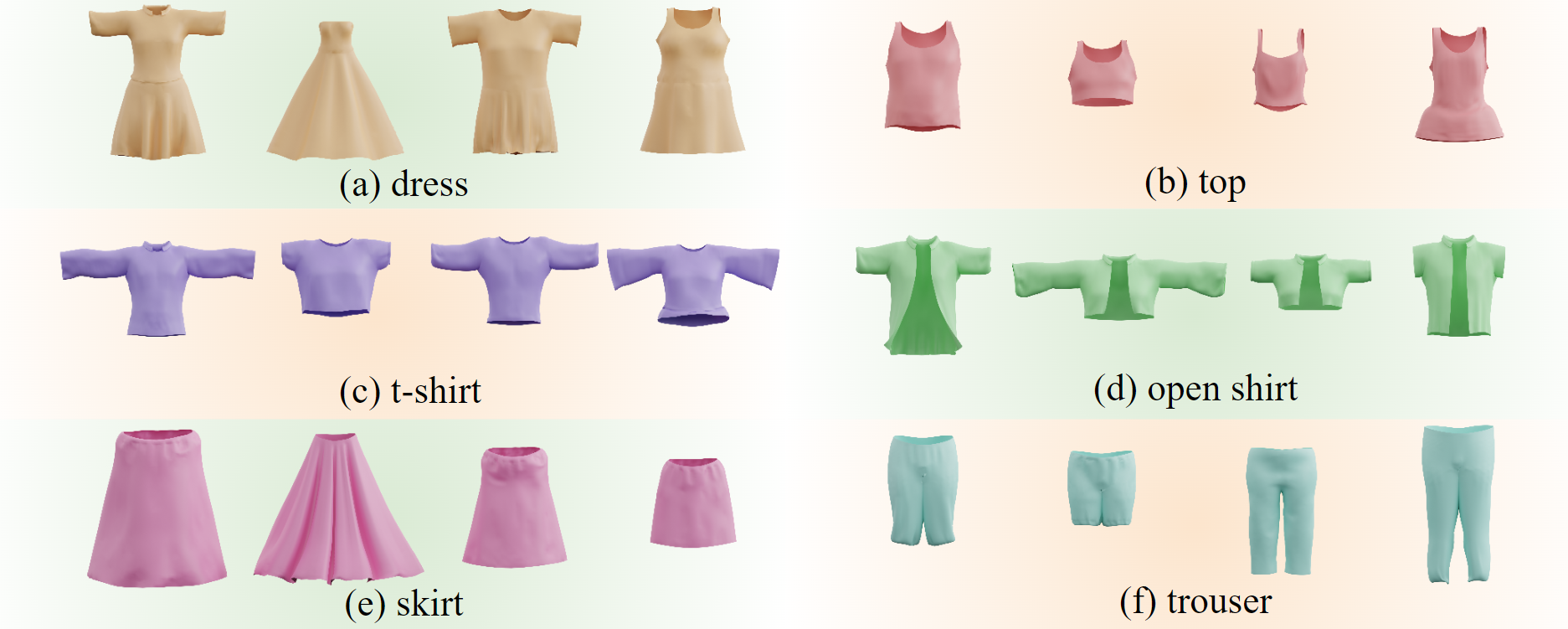}
  \caption{Dataset examples of (a) dresses, (b) tops, (c) t-shirts, (d) open shirts, (e) skirts, and (f) trousers. }
  \label{fig:dataset}
\end{figure}

To train and test our framework, we leveraged the public synthetic 3D clothes dataset, Cloth3D++ \cite{cloth3d, cloth3d++}. 
The dataset encompasses a variety of garment types, including dresses, tops, t-shirts, open shirts, skirts, and trousers, examples of which are showcased in Figure \ref{fig:dataset}.
In total, we used six types of garments for our training process, incorporating 4300 garments, and we reserved an additional 100 garments for quantitative experiments in validation. 
Note that the garments for evaluation were not included in the training set.
Additionally, we employed some real-world 3D scanned garment data for qualitative  results.
All garments were roughly aligned to a mean shape, T-posed body (zero-body). 
We checked if the garment in the rest state is located below a specific threshold with respect to the zero-body (which is set to around the collarbone), if then, we applied the holding loss during the training phase. 
We employed the AMASS \cite{AMASS} dataset in which random poses were sampled for training, and shape parameters were randomly sampled in the range of [-2,2] for all elements.
We use 300 shape components for the body model \cite{SMPL-X}.

\subsection{Implementation Details}

We conducted all experiments in this paper using an NVIDIA 3090 GPU, 128Gb RAM, and an Intel i7-10700 CPU. 
For the clothes embedding network, which is implemented with DiffusionNet \cite{diffusionnet}, we utilized 8 DiffusionNet blocks and the width of which are set to 512. 
We did not perform normalization for input position nor augmentation, and other setups follow the original paper. 
The dimensions of the clothes embedding, $z_v$, the single draping embedding, $z_s$, and the force vector between garments $f^{ij}$ are set to 512.
The single draping embedding network $\mathrm{F_s}$ and force embedding network $\mathrm{F_f}$ are implemented with six-layered MLPs with a hidden dimension size of 512. 
The intermediate activation function is GeLU \cite{gelu}, and a residual connection connects every two layers. 
The skinning network is implemented with a MLP that consists of 4 hidden layers with a feature size of 128. 
The final output value of the skinning network is multiplied by 0.01, and the rest of the settings are the same as other MLP networks.
The per-vertex displacement prediction networks, $\mathrm{D_s}$ and $\mathrm{D_m}$, are implemented as modulated SIREN \cite{modulatedSIREN}, and both the SIREN network and the modulator have six hidden layers with a feature size of 512. The outputs of $\mathrm{D_s}$ are multiplied by 0.1, and the outputs of $\mathrm{D_m}$ are multiplied by 0.01. 
Learning was conducted for $200,000$ iterations with a batch size of 1, and progressive learning was performed from T-pose to full pose until $100,000$ iterations where the entire training time took about three days. The learning rate was set to $0.0001$, the ADAM optimizer was used, and the gradient clipping was performed by setting the maximum norm to 1. 
Our untangling network is designed as a graph neural network which is invariant to number of nodes, and since we desire to generalize to arbitrary number of layers, we randomly set the number of layers $M\in[1,6]$, and also randomize their order during training.
Additionally, an automatic post-processing step to resolve minor penetrations by comparing vertices' positions and normals between layers was applied to compensate for slight penetrations, as is common in previous works \cite{GarNet, snug, santesteban2019, TailorNet}, in the qualitative evaluation.

\begin{figure}[t]
  \includegraphics[width=1.0\linewidth]{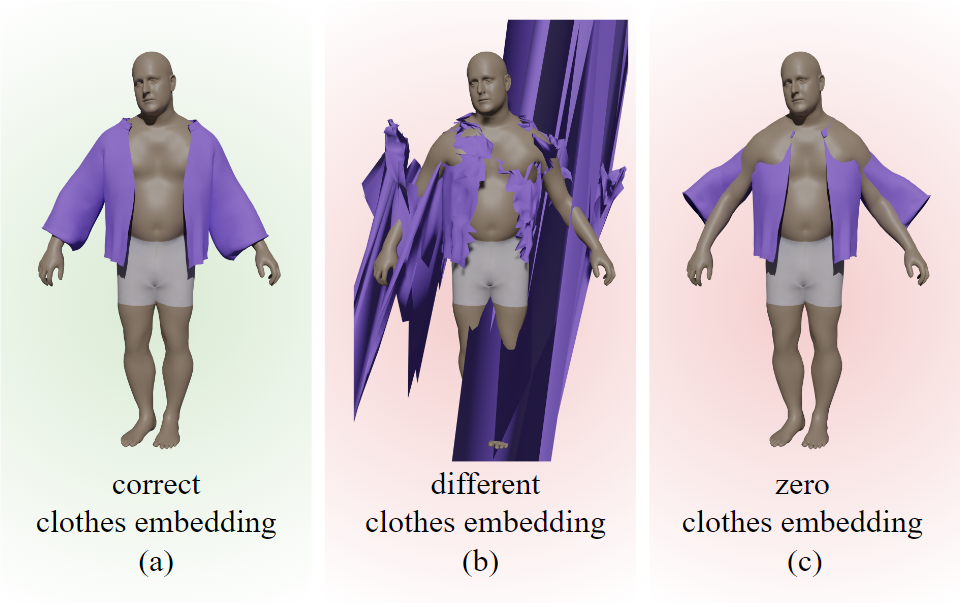}
  \caption{The results of using either correct or incorrect embedding in the single draping network: (a) correct embedding, (b) embedding derived from different clothes, (c) embedding of which elements are set to zeros. }
  \label{fig:wrong_embeddings}
\end{figure}

\begin{table}[t]
\caption{ Comparison of the losses between when the correct clothes embedding is provided (correct) and when the incorrect clothed embedding is provided (different, zero).
}
\label{table:wrong_embedding}
\vskip 0.1in
\begin{center}
\begin{small}
\begin{sc}
\begin{tabular}{lccccr}
\toprule
\#Layer & Correct & Different & Zero \\
\midrule
Strain &0.899 &$2.77 \times 10^{14}$ &$1.79\times 10^{2}$ \\
Bending &0.081 &$5.50\times 10^{5}$ &0.116\\
Collision &0.570 &$1.09\times 10^{2}$ &$1.87\times 10^{2}$ \\
Gravity &-0.623 &-0.553 &-0.570 \\
\bottomrule
\end{tabular}
\end{sc}
\end{small}
\end{center}
\vskip -0.1in
\end{table}

\begin{figure*}[t]
  \includegraphics[width=1.0\linewidth]{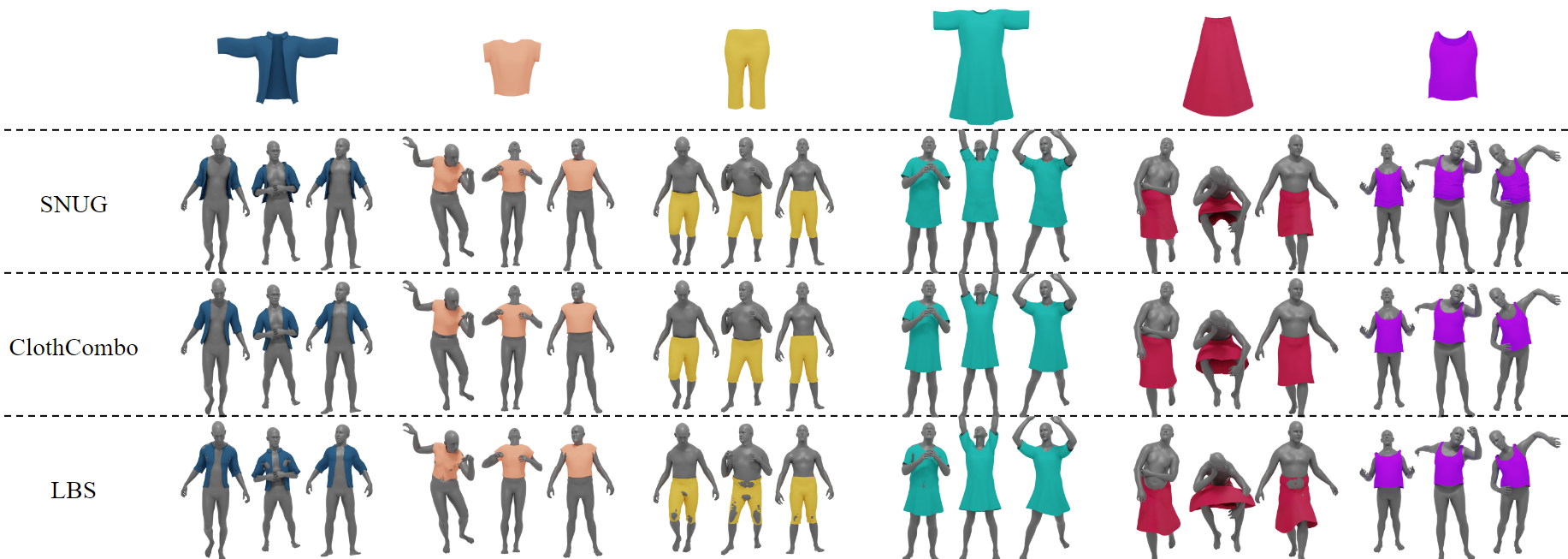}
  \caption{single garment draping results comparison examples}
  \label{fig:vssnug}
\end{figure*}

\begin{table}[t]
\caption{
Comparison of the average energy terms between SNUG, our single clothing draping network, and LBS baseline in six different garment categories. 
}
\label{table:vsSNUG}
\vskip 0.1in
\begin{center}
\begin{small}
\begin{sc}
\begin{tabular}{lcccccr}
\toprule
Metric & Strain & Bending & Gravity & Collision\\
\midrule
SNUG &\textbf{0.737} &0.115 &-0.460 &\textbf{0.165} \\
Ours &0.955 &0.105 &\textbf{-0.478} &0.620    \\
LBS &2.177 &\textbf{0.072} &-0.463 &222.521 \\
\bottomrule
\end{tabular}
\end{sc}
\end{small}
\end{center}
\vskip -0.1in
\end{table}

\subsection{Clothes Embedding}

Our proposed method incorporates the clothes embedding network, which converts garments of any topological structure into a condensed and expressive clothes embedding, represented by $z_v$. 
To evaluate the efficacy of our network, we conducted an experimental assessment in which we analyzed the result of the single clothing draping network with correct clothes embedding and incorrect clothes embeddings.
The incorrect clothes embeddings are generated in two ways: by using the clothes embedding of different garments or setting the clothes embedding to zero.
The results, depicted in Figure \ref{fig:wrong_embeddings}, demonstrate that when the correct embedding was used in the single clothing draping network, the garment was accurately draped over the target body. Conversely, when incorrect garment states were used (Figure \ref{fig:wrong_embeddings}(b), (c)), the simulated garments failed to fit to the body correctly, especially when different embedding was used. 
We quantitatively compared these results in Table \ref{table:wrong_embedding}. It shows a significant increase in strain, bending and collision losses when incorrect clothes embedding is used.
This comparison verifies that our clothes embedding network effectively encodes the garment information. 

\begin{figure*}[t]
  \includegraphics[width=1.0\linewidth]{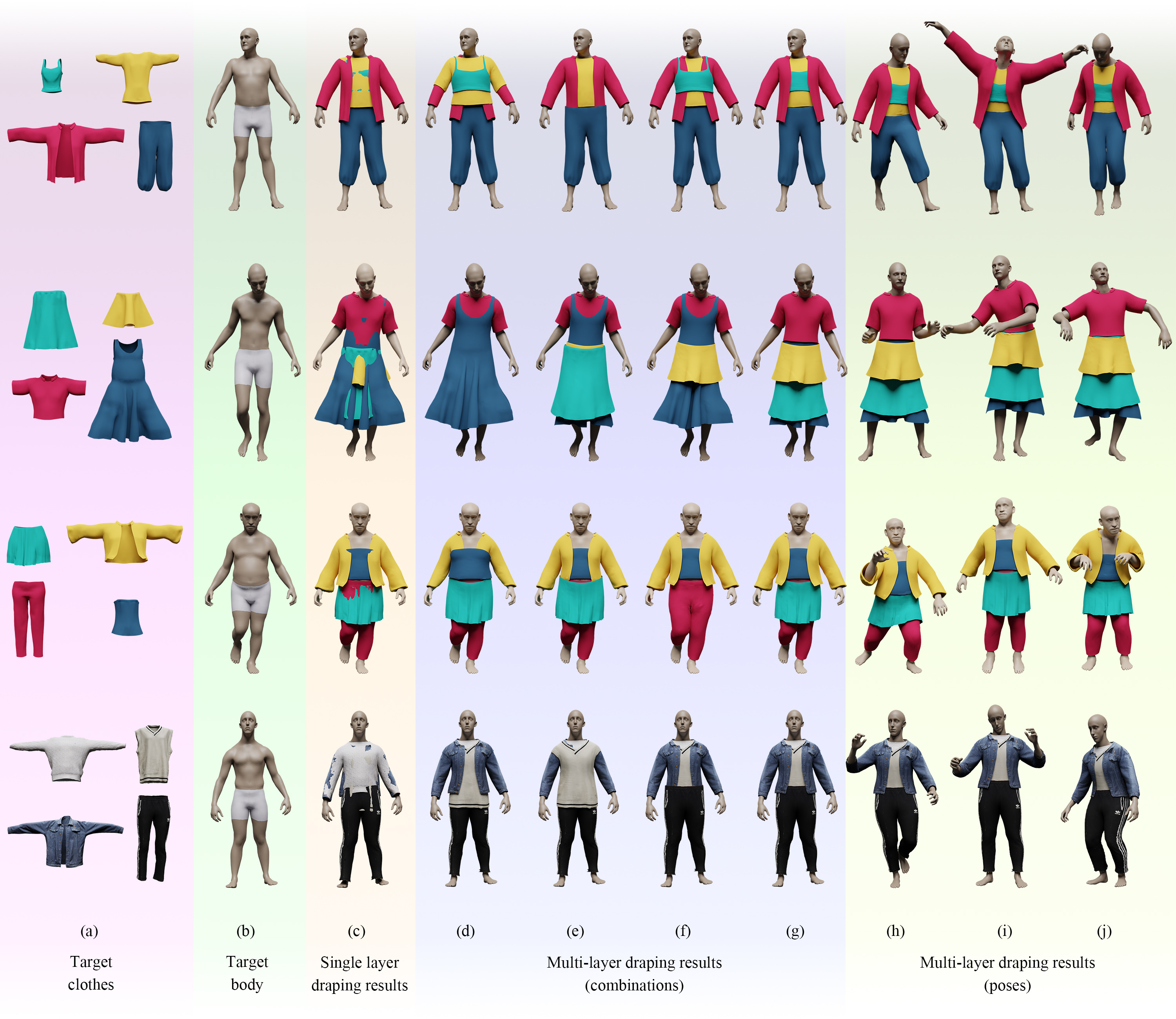}
  \caption{Results of draping various garments in different orders and combinations onto diverse body shapes and poses. 
  Each row presents results with the same set of garments and the target body.
  The first three rows show the results derived from synthetic Cloth3D++ dataset, and the last row shows results using scanned data of real-world garments. 
  }
  \label{fig:result1}
\end{figure*}

\begin{figure*}
  \includegraphics[width=1.0\linewidth]{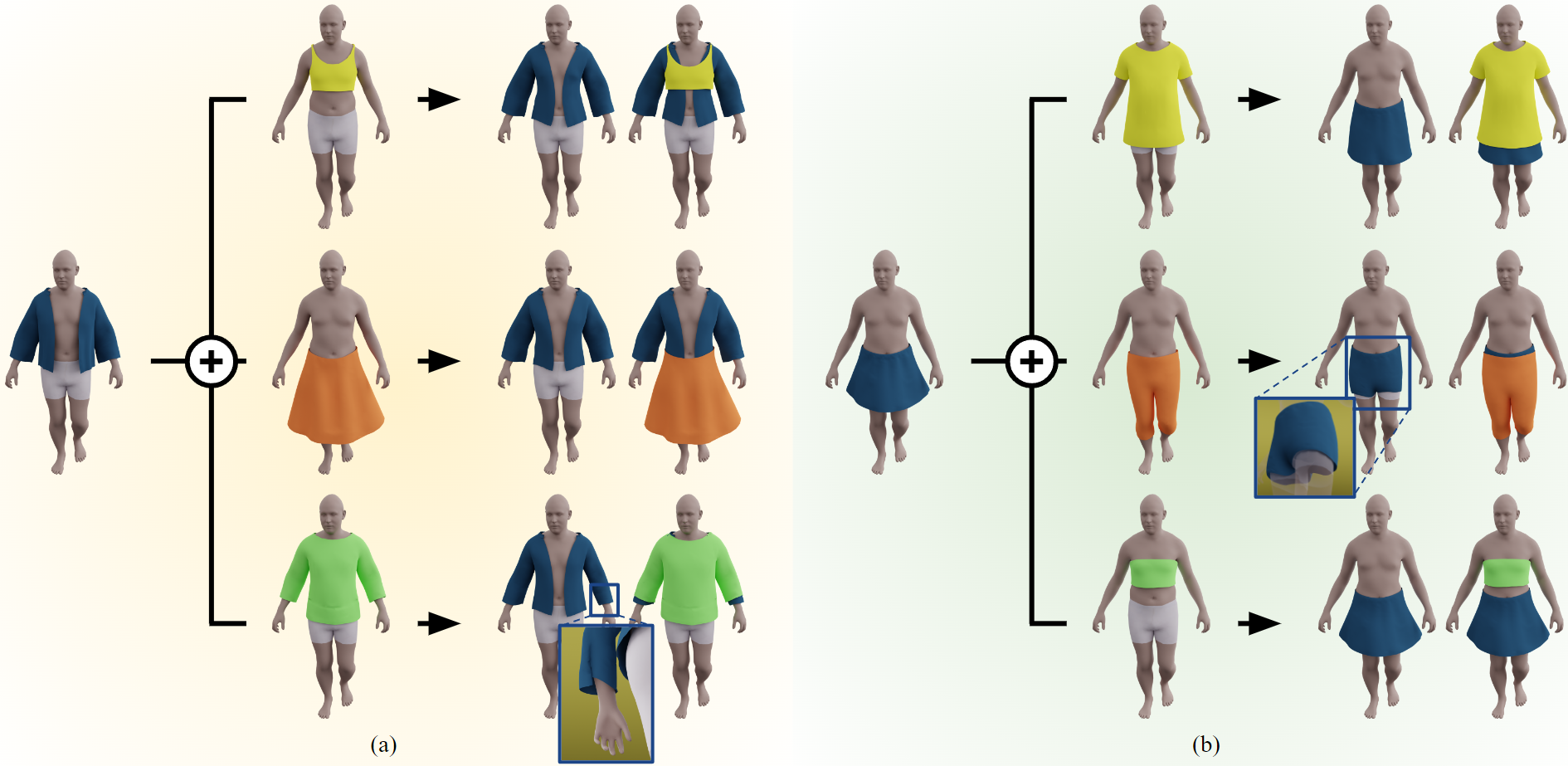}
  \caption{
  Results of layering different clothes on the same garment: (a) a top, a skirt, a long-sleeved T-shirt on an open jacket, (b) a T-shirt, a trouser, a top on a skirt. Notice how each clothes deforms differently with respect to the clothes that are dressed on top.
  }
  \label{fig:layer_deform}
\end{figure*}

\subsection{Single Clothing Draping}
The proposed single clothing draping network enables the dressing of garments with arbitrary topologies onto bodies of various shapes and poses. This is achieved by leveraging a unified clothes embedding, which ensures a robust representation of different garment types within the same latent space.
To evaluate the performance of our single clothing draping network, we compare it with the garment-specific model, SNUG \cite{snug}
, which is generally on par or outperforms other draping methods such as PNBS \cite{PBNS} in most previous works.
We trained SNUG for each garment of six different types and compared the results with our single clothing draping network and the baseline method, LBS.
The clothes' skinning weights used in SNUG and LBS were assigned based on the nearest body vertex and refined using Laplacian smoothing so that vertices in proximity have similar skinning weights. 
Table \ref{table:vsSNUG} presents the average strain, bending, gravity, and collision loss for each method validated with 500 random shapes and poses.
While the garment-specific model outperforms ours in several metrics such as strain and collision energies, our single clothing draping network can still preserves the physical constrains better while achieving realistic draping compared to the LBS baseline. 
Figure \ref{fig:vssnug} provides qualitative results demonstrating this interpretation.
Our proposed method achieves more physically plausible draping results compared to the LBS baseline, although they may exhibit fewer detailed wrinkles compared to SNUG. 
However, considering that our method does not require garment-specific training and can be applied to any type of garments, it is a reasonable trade-off. 
Furthermore, while our network can be extended to the draping of multi-layered garments, it is more challenging to extend garment-specific models to multi-layered settings, as this requires training for every combination of clothes.

\subsection{Multi-Layered Clothes Draping}

To qualitatively evaluate our draping results, we trained a model with a maximum $M=6$. During training, the number of layers $M$ was randomly set between 1 to 6 for each iteration.
Figure \ref{fig:result1} illustrates the examples of draping results with four layers.
The figure shows the target garments in (a), the target body in (b), and the overlapped single clothing draping results in (c).
The columns (d-g) display the results of multi-layered draping results in different orders of clothes, and those in (h-j) show the posed result of the avatar in (g).
The figure shows that our model correctly drapes the target garments in complex layering scenarios.
For instance, in the first row, the blue pants could be draped either on top of other clothes (d) or below (e).
In the second row, in contrast to (c), in which the blue dress exhibits severe inter-penetrations with other clothes, it is properly deformed to be draped within other garments, as shown in (d-g).
The draping results are also plausible in diverse poses, as can be seen in (h-j).
Note that all clothes have different topologies, demonstrating the ability of our model to generalize well to garments with various topologies. The last row of the figure presents the results of draping the real 3D-scanned clothes data, which have less organized topologies than the synthetic data. The results show that our model is still capable of handling real clothes while producing plausible draping results. 

To further evaluate the influence of layering different types of clothes, we conducted experiments where different garments were draped over the same base garment. Figure 7 (a) illustrates the results of draping three different garments over an open blue shirt. When the yellow top bra is draped on top, only the upper part of the shirt is squashed inward. In contrast, when the green t-shirt is draped, it exhibits proper folding and shrinking in the sleeve part. In (b), we draped three different garments over a blue skirt. When the yellow T-shirt is draped on top, the blue skirt is pressed down by the T-shirt. Likewise, the skirt undergoes severe deformation clinging to the lower body when we draped the orange pants. 
Conversely, if a non-contact garment is draped on top, as in the case of the green top, no deformation occurs. Additionally, in both (a) and (b), the garments in the upper layers exhibit slight expansion due to the influence of the underlying clothes, which is physically plausible. The green top in (b) is an exception, as the skirt does not influence the top. This demonstrates that our untangling model appropriately models the interactions between garments.

\begin{figure*}
  \includegraphics[width=1.0\linewidth]{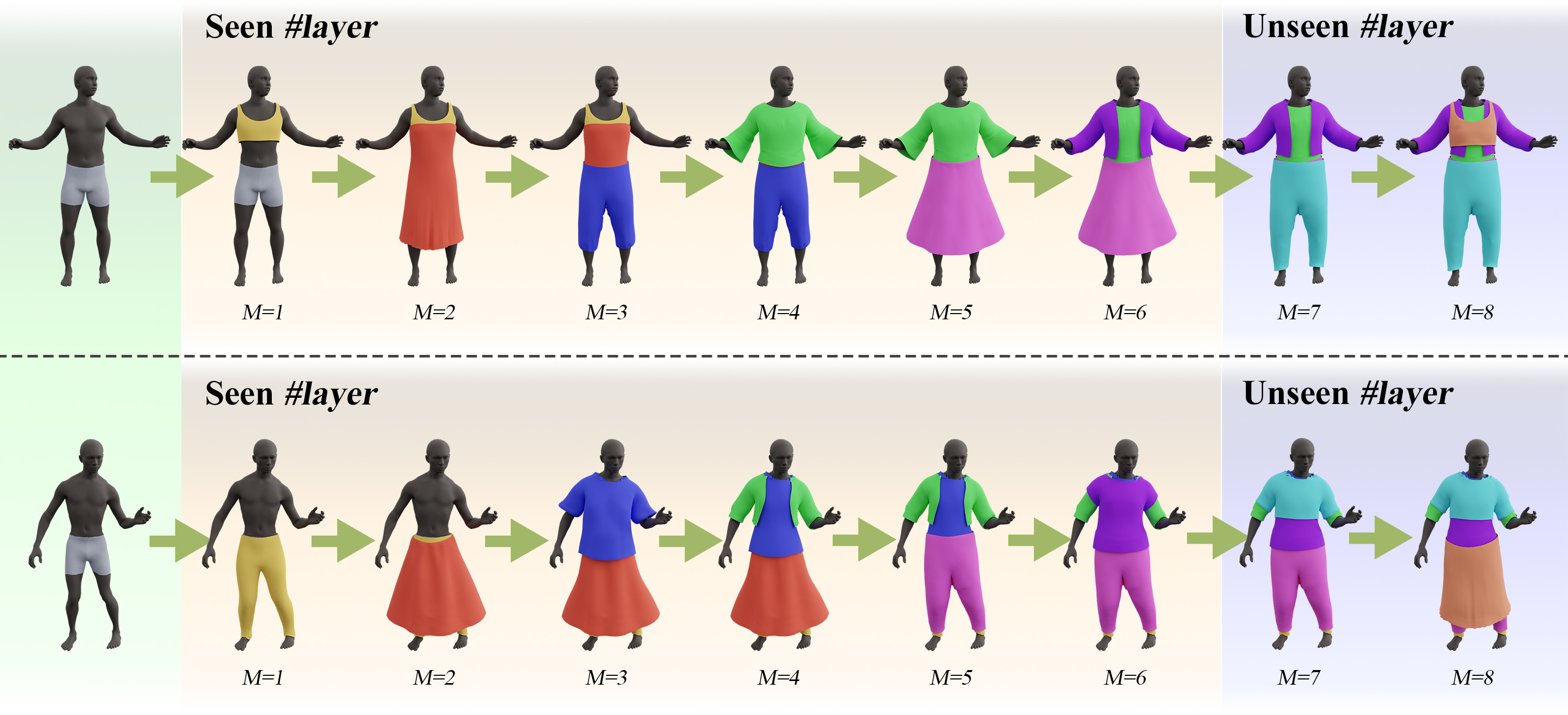}
  \caption{
  Results of sequentially layering clothes. The network trained to drape a maximum of six layers, not only generalizes well to layer configurations ranging from 1 to 6 clothes, but also 7 and 8 clothes. 
  }
  \label{fig:multilayers}
\end{figure*}

Our untangling model, based on a GNN, is invariant to the number of nodes. This allows it to accommodate an arbitrary number of garment layers. Figure \ref{fig:multilayers} provides two examples of sequential garment layering, demonstrating our method's capability to dress any number of layers in a physically plausible manner. Specifically, the purple section presents the results of draping unseen numbers of layers during training. 

Untangling intricately tangled garments is a complex problem, making it difficult to obtain ground-truth physically-based simulation results. Therefore, instead of directly comparing the results with ground-truth counterparts, we have conducted extensive evaluations on physics based energy terms.
Firstly, we performed a numerical analysis to gauge the degree of deformation of each layer for various numbers of layers and have presented the results in Table \ref{table:multilayers}. The values in the table represent the ratio of strain energy resulting from multi-layered draping to the strain energy resulting from single clothing draping. Each row corresponds to the total number of layers, $M$, while each column represents the index of the layer ($1$ for the innermost). 
As the table illustrates, a lower number of layers minimizes the overall increase in strain energy due to interaction between garments. 
However, as the number of layers increases, so does the strain energy. 
This indicates that adding more layers of clothing leads to more significant stretching or shrinking to prevent collisions.
This inference aligns with the real-world observation that garments tend to stretch or shrink more due to the pressure from their interaction when multiple layers are worn.
Additionally, the outermost garment layers experience the largest increase in strain energy, as they are compressed by all other layers. 
Taken together, the qualitative results demonstrate our model's ability to realistically layer multiple layers of clothing, while the quantitative results show that the deformation experienced by each clothing layer is physically reasonable.
This experiment demonstrates the robustness of our model in handling various numbers of garments.

\begin{figure}[t]
  \includegraphics[width=1.0\linewidth]{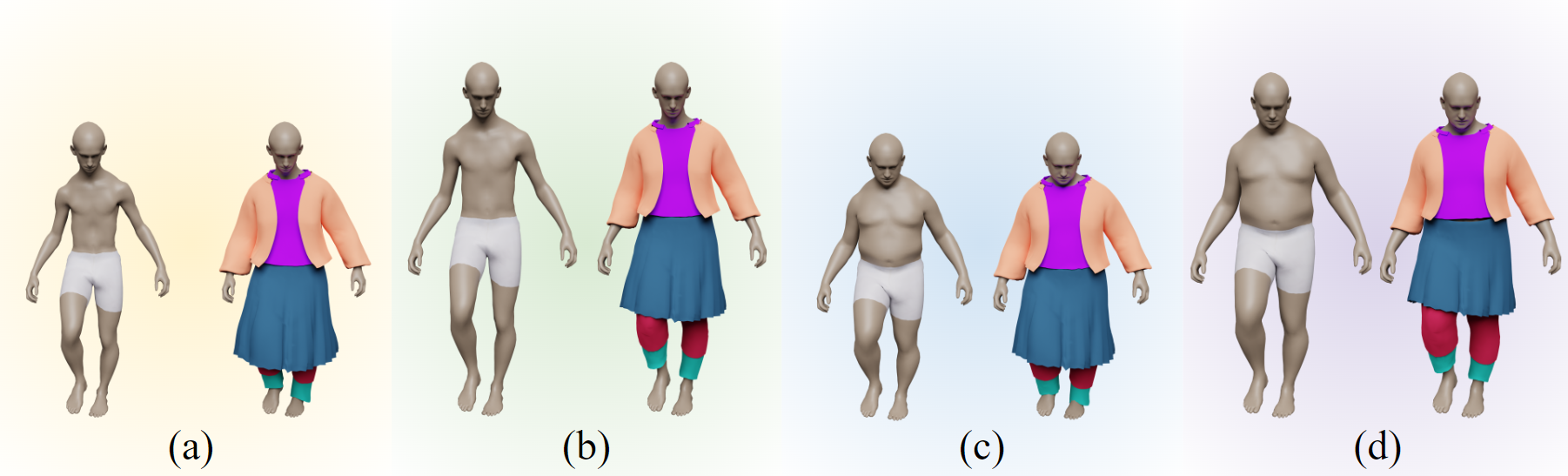}
  \caption{Results of draping the same set of garments on diverse body shapes. Each column, distinguished by different colors, presents a pairing of a specific body shape (on the left) and the corresponding draping outcome (on the right). 
  }
  \label{fig:multishape}
\end{figure}

\begin{table}[t]
\caption{
Ratio of strain loss in the multi-layered draping result to that in the single clothing draping result in each layer.
}
\label{table:multilayers}
\vskip 0.1in
\begin{center}
\begin{small}
\begin{sc}
\begin{tabular}{lcccccccr}
\toprule
\#Layer & $1$ & $2$ & $3$ & $4$ & $5$ & $6$ \\
\midrule
$M=1$ &1.00 &- &- &- &- &-  \\
$M=2$ &1.10 &1.29 &- &- &- &-  \\
$M=3$ &1.16 &1.28 &1.33 &- &- &-  \\
$M=4$ &1.32 &1.47 &1.43 &1.83 &- &-  \\
$M=5$ &1.40 &1.63 &1.45 &1.69 &2.07 &-  \\
$M=6$ &1.61 &1.46 &1.44 &1.63 &1.72 &2.65  \\
\bottomrule
\end{tabular}
\end{sc}
\end{small}
\end{center}
\vskip -0.1in
\end{table}

Figure \ref{fig:multishape} illustrates the generalizability of our model to different body shapes.
We draped the same combination of clothes, with $M=5$, onto various body shapes, ranging from short to tall and thin to fat. In each case, the clothes are draped in a physically accurate manner. For instance, the sleeves of the top garments appear shorter on taller bodies (b, d) compared to those on shorter bodies (a,c).

\begin{figure}[t]
  \includegraphics[width=1.0\linewidth]{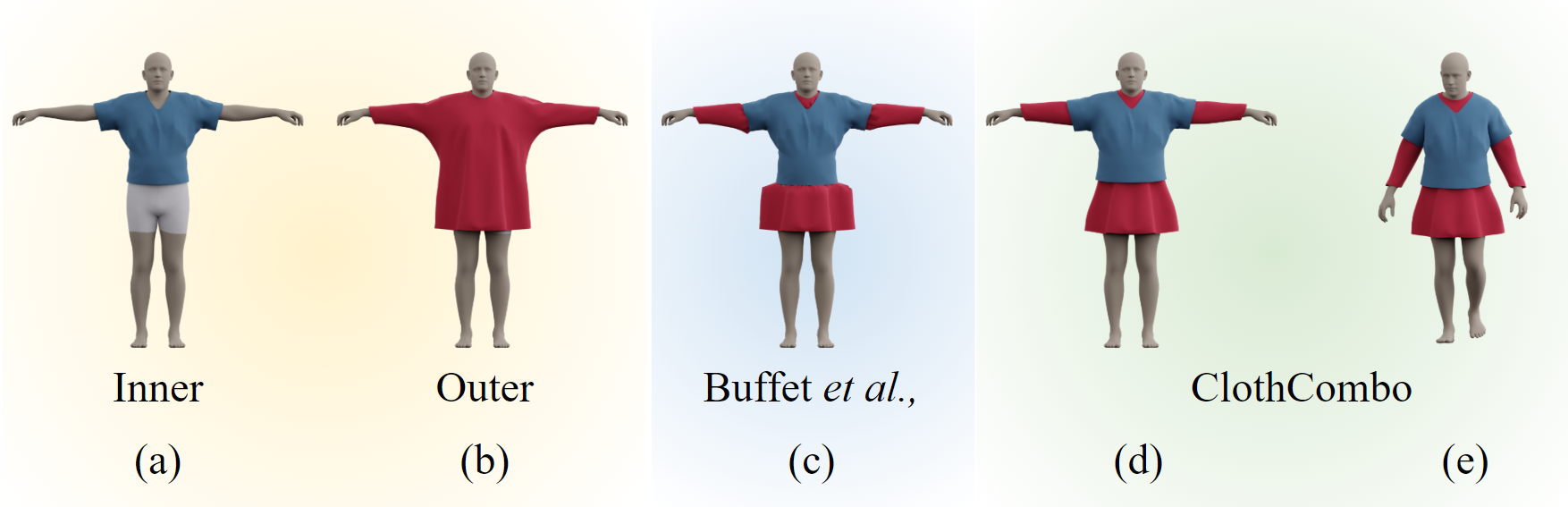}
  \caption{Comparison of the results of draping two-layer garments.}
  \label{fig:comparison}
\end{figure}

Although a garment-specific pre-training process is required, the untangling component of ULNeF \cite{ULNeF} can be applied to unseen garments. 
The neural untangling network in ULNeF is trained using ground-truth data generated by the untangling operator \cite{implicitUntangling}. 
Therefore, we compared our results of multi-layered draping with the untangling operator \cite{implicitUntangling}.
In Figure \ref{fig:comparison}, (a) and (b) are the results of our single clothing draping network, and (c) and (d) show the multi-layered garment draping results obtained using the untangling operator\cite{implicitUntangling} and ClothCombo, respectively.
As seen in (c), when the lower-layer garment (a) is much larger than the upper-layer garment (b), the untangling operator struggles particularly at the border of the garment. 
This is because the untangling operator is locally applied to the area where penetration occurs, and cannot consider outside of the covariant field. This results in severely distorted results in complex multi-layered scenarios.
On the other hand, ClothCombo is able to deform the garments naturally as in (d) and, unlike ULNeF, it also enables posed simulations as in (e).

\subsection{Ablation Study}

\begin{figure}[t]
  \includegraphics[width=1.0\linewidth]{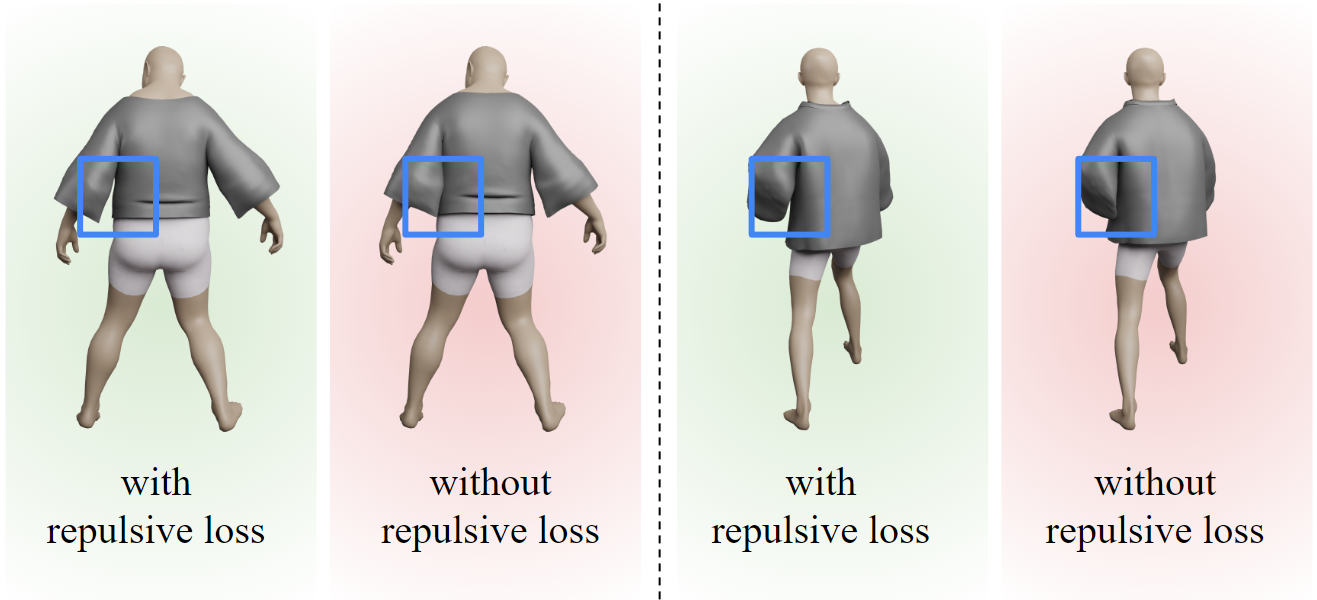}
  \caption{Results generated with and without the repulsive loss term in the case of two different poses. 
  }
  \label{fig:repulsive}
\end{figure}

\begin{figure}[t]
  \includegraphics[width=1.0\linewidth]{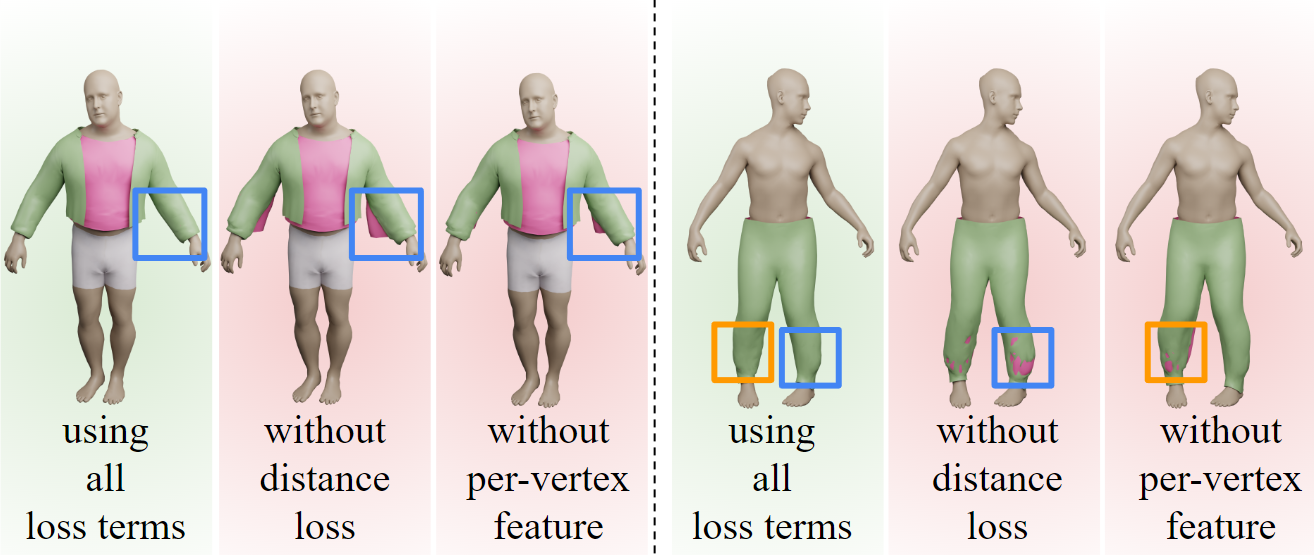}
  \caption{Results generated when excluding various loss terms.}
  \label{fig:loss_ablation}
\end{figure}

\begin{figure}[t]
  \includegraphics[width=1.0\linewidth]{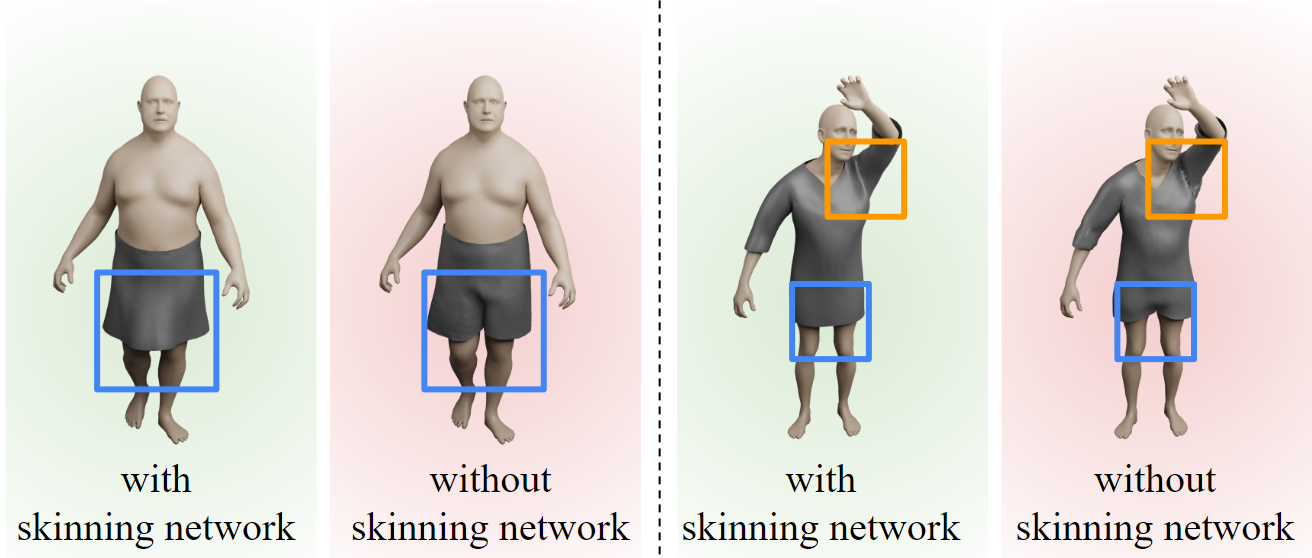}
  \caption{Results generated with and without skinning weight optimization.
  }
  \label{fig:skinning}
\end{figure}

We evaluated the efficacy of our loss terms by omitting each and that of the per-vertex clothes embedding ($z_v$) by replacing it with the global clothes embedding ($\bar{z}_v$). We use strain loss, bending loss, collision loss between garment and body, and multi-collision loss between garments as metrics for comparison.
The degree of self-collision is also measured to assess the amount of self-interpenetration, similar to the repulsive loss, but applied only to the vertices where penetration occurs. 
As demonstrated in Table \ref{table:ablation}, omitting the repulsive loss term ($L_r$) does not result in a significant difference in physical metrics, except for an increase in self-collisions. This is clearly demonstrated in Figure \ref{fig:repulsive}, where there is a noticeable increase in self-interpenetration, particularly in the blue box area. Removing the distance loss term ($L_d$) or per-vertex feature ($z_v$) leads to a significant increase in multi-collision loss ($collision_{gg}$). 
As shown in the figure \ref{fig:loss_ablation}, removing the distance loss term or per-vertex feature may result in penetration between garments which is not physically permissible.
Similarly, omitting the optimization for skinning weights also results in larger errors with distortions. This is evident from the blue and orange boxes in Figure \ref{fig:skinning}.

\begin{table}[t]
\caption{Ablation Study: A comparison of various garment states when each loss term is not utilized. Beginning from the second column, each column represents the case of using all losses (All), omitting the repulsive loss ($L_r$), omitting the distance loss ($L_d$), removing the per-vertex feature ($z_v$), and omitting the skinning network ($\mathrm{S}$) where the skinning network is inactivated, in that order.
}
\label{table:ablation}
\vskip 0.1in
\begin{center}
\begin{small}
\begin{sc}
\begin{tabular}{lccccccr}
\toprule
Metric & All & w/o $L_r$ & w/o $L_d$ & w/o $z_v$ & w/o $\mathrm{S}$ \\
\midrule
strain$\downarrow$ &0.985 &0.978 &\textbf{0.967} &1.115 &3.430    \\
bending$\downarrow$ &0.103 &0.099 &0.097 &\textbf{0.093} &0.181    \\
collision$_{gb}$$\downarrow$ &\textbf{0.409} &0.412 &\textbf{0.409} &0.413 &0.901    \\
collision$_{gg}$$\downarrow$ &\textbf{0.138} &0.155 &0.309 &0.377 &0.312    \\
self-collision$\downarrow$ &0.779 &0.784 &\textbf{0.778} &0.792 &0.922    \\
\bottomrule
\end{tabular}
\end{sc}
\end{small}
\end{center}
\vskip -0.1in
\end{table}

\begin{table}[t]
\caption{The per-garment inference time for each phase of the proposed method. Note that the garment embedding step is only required to be performed once for each garment. The number of vertices per garment is 8K on average.
}
\label{table:runtime}
\vskip 0.1in
\begin{center}
\begin{small}
\begin{sc}
\begin{tabular}{cccccr}
\toprule
Clothes & Single Clothing & Untangling & Post \\
Embedding & Draping Network & Network & processing \\
\midrule
27 ms &0.6 ms &0.4 ms &200$\sim$300 ms    \\
\bottomrule
\end{tabular}
\end{sc}
\end{small}
\end{center}
\vskip -0.1in
\end{table}

\section{Discussion and Conclusion}
\label{sec:conclusions}

In this paper, we introduce a novel method called ClothCombo for draping arbitrary combinations of clothes on diverse body shapes and poses. While effectively addressing highly complex inter-cloth interactions in draping multi-layered clothes, our method has shown the generalization capability to body shape, pose, clothes topology, and the number of clothes. We extensively evaluate our approach in complex multi-layered clothing scenarios and through ablation studies, validate how each component is essential to result in a collision-free state. We also demonstrate the applicability of our method in layered virtual try-on of real clothes. 

Although our method has demonstrated appealing results, it has several limitations we plan to address in the future. First, the resulting geometry of each clothes is smooth with few details such as wrinkles and folds. This may be due to the clothes embedding network that fails to capture local features. Locally processing clothes as in \cite{hood, physgraph} can be considered in the future. Second, our method sometimes fails to perform in extreme cases, such as draping tight shorts on top of long skirts. 
Third, the secondary dynamics are not considered as our network is conditioned solely on current pose. Unlike SNUG\cite{snug} and PBNS\cite{PBNS}, our draping network does not capture the detailed wrinkles effectively, especially in the case of long dresses or skirts. One way to address this issue would be utilizing both static encoder and dynamic encoder for modeling pose-dependent deformation and motion-dependent deformation respectively as in \cite{bertiche2022neural}.
Lastly, although our loss terms effectively prevent penetrations between clothes, there can still be intersections during the inference phase. 
This can be addressed by performing an additional post processing step as did in most previous works \cite{GarNet, snug, santesteban2019, TailorNet}. However, this limits the deployment of our method to real-time applications. Although our network’s inference can be made in real-time, the post processing step can be time-consuming (see Table \ref{table:runtime}). It would be valuable for future work to address this issue by restricting the intersections between garments more effectively at the network level. One possibility is using the concept of diffused human body model proposed in \cite{santeteban2021}, which significantly reduces the body-clothes intersections. Other possible future works may include exploring the dynamics of clothing and incorporating physical characteristics of clothes into the model.

\begin{acks}
This research was supported by the National Research Foundation of Korea(NRF) grant funded by the Korea government(MSIT). (No. NRF-2020R1A2C2014622)
\end{acks}

\bibliographystyle{ACM-Reference-Format}
\bibliography{sample-bibliography}

\end{document}